# The Network of French Legal Codes

Pierre MAZZEGA
Université de Toulouse; UPS(OMP)
CNRS IRD, LMTG Toulouse
14 av. Belin 31400 Toulouse, France
+33 (0)5 61 33 25 64
mazzega@lmtg.obs-mip.fr

Danièle BOURCIER
CERSA CNRS
Université de Paris 2
10 rue Thénard 75005 Paris
+33(0)1 42 34 58 96
daniele.bourcier@cersa.cnrs.fr

Romain BOULET
Université de Toulouse; UPS(OMP)
CNRS IRD, LMTG Toulouse
14 av. Belin 31400 Toulouse, France
+33 (0)5 61 33 25 02
boulet@lmtg.obs-mip.fr

## ABSTRACT
We propose an analysis of the codified Law of France as a structured system. Fifty two legal codes are selected on the basis of explicit legal criteria and considered as vertices with their mutual quotations forming the edges in a network which properties are analyzed relying on graph theory. We find that a group of 10 codes are simultaneously the most *citing* and the most *cited* by other codes, and are also strongly connected together so forming a "rich club" sub-graph. Three other code communities are also found that somewhat partition the legal field is distinct thematic sub-domains. The legal interpretation of this partition is opening new untraditional lines of research. We also conjecture that many legal systems are forming such new kind of networks that share some properties in common with small worlds but are far denser. We propose to call "*concentrated world*".

## 1. INTRODUCTION
Local, national, regional and international Laws are experiencing major transformations within the stream of globalization and intensification of the social, technical, economical and legal interactions through all the scales of the societal organization. This transformation can be observed on at least one of its phenomenological dimension: the hierarchies of legal systems are increasingly embedded in network-like structures [12] that link together in closer neighborhoods different legal sub-systems, legal fields and authorities active at various governance levels [10]. We here propose to have a bird-eye view of the whole French legal system, focusing on a particular structural property: the networking of the French legal codes (hereafter FLC for short). Such a global view opens new perspectives in the scientific characterization and measuring of "*networking*" dimension of the legal complexity [4].

## 2. BUILDING THE *FLC* NETWORK
Our approach is made possible because of the launching of a policy of codification in France since several decades. In 1989, the *Commission Supérieure de Codification* has been created near the Prime Minister. Its missions are first to gather, organize and link all the existing codes with the scattered laws and decrees, and then to build new thematic and comprehensive codes on the ground of the existing laws in the domain of interest. In 2009 at least 60 % of the legal texts have been codified following this method. A noticeable exception is the tax law that still does not figure out in the list of legal codes due to the specific way used by the French *Ministère du Budget* (Ministry of Budget) to build the Tax Code. In this study we only consider the French codes that have actually been built following the same methodology and with the same criterion elaborated and published by the *Commission de Codification*. This set of codes is representative of the current state of the legal corpus in France (see the website of LEGIFRANCE [8]).

As we are mainly interested in the structural properties of the FLC network, we search in the text of each code which other codes are cited. We do not care about the numerous self-quotations. We also do not mind whether a whole code is cited *per se*, or just one of its constitutive book, title, chapter, or article. In other words the only objects that we recognize are codes and we do not consider the internal organization of the codes (as we have done in [5] looking for other properties of legal networks at finer scales). Applying these rules for the selection of codes, we obtain a set of 52 validated codes. With these data we establish a link from code *X* to code *Y* when *X* is citing code *Y* at least one time (we say that *Y* is the *cited* code, *X* the *citing* code). In the graph codes *X* and *Y* are vertices, and the link is the edge from *X* to *Y*. The resulting graph is drawn on the Figure. It is dense and complex so that its analysis must rely on statistical concepts and tools [6].

## 3. SINGULAR CODES

The graph exhibits a single isolated vertex (Code of the Honor Legion and Military Medals), a pendant vertex (with single link: Code of the Monetary Instruments) and a code citing four codes but never cited itself (Code of the Handicraft). The **most citing** codes (yellow squares and cyan hexagon on the figure) are the general code of the local and regional Authorities, the environmental code, the code of the public health, the code of the social security and the rural code. These codes are covering a large diversity of matters. The six **most cited** codes (green circles and cyan hexagon on the figure) are the penal code, the civil code, the code of the penal procedures, the code of labor, the code of commerce and the code of public health. The penal code is most cited because it gathers all the penalties associated to any kind of offense or infringement of a legal nature. Two reasons are susceptible of explaining the second rank found for the civil code: a) it manages a large corpus of norms related to many domains of the private law (the persons, property, securities); b) it is directly inspired from the oldest code in France, dating back from Bonaparte's consulate (1804).

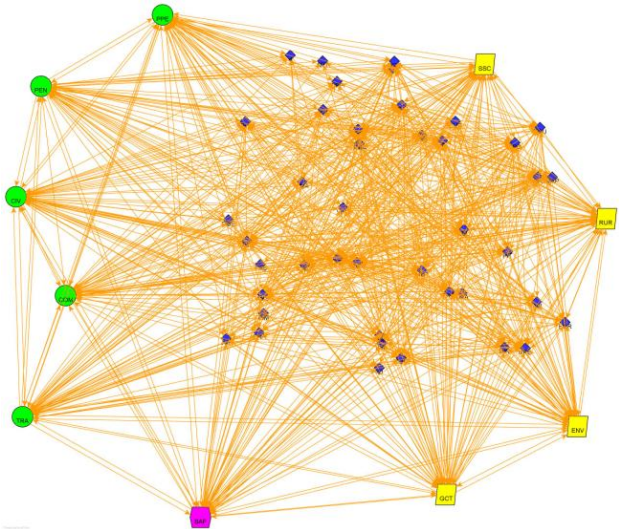

Figure. The graph associated with the network of French Legal Codes. Each vertex (blue diamonds, yellow squares, green circles, cyan hexagon) represents a Code. The yellow squares stand for the 4 codes most citing other codes. The 5 green circles are the most cited codes. The cyan hexagon is in both sets of the most cited and most citing codes. All these codes together (not including the diamond shaped symbols) form the rich club.

Using some mathematical definition of centrality in use in the analysis of social networks [1, 2], we show that these 10 codes are the most central in the FLC network: they both capture a high percentage of the quotations and are strongly linked between each other, so forming a "*rich club*". Then discarding these 10 codes, we can partition the resulting reduced network in three main components (sub-graphs) that present a high degree of internal connectivity (using objective statistical criteria [7]). These three code communities are somewhat partitioning the legal field: one is concerned with issues related to territories, resources and property (13 codes); the second one is about regulations of social systems and activities (12 codes); the third one mainly deals with administrations and related domains (12 codes). However this provisional interpretation of the observed partition is opening new untraditional lines of research. Interestingly, we also show that the FLC network is much denser than most of the social networks [9, 11]. We conjecture that such combined property of being dense and presenting some central community (the rich club), is a specific feature of many legal systems seen at this scale (at least the national level) and resolution (codes or laws are the smaller object that we consider). We propose to call "*concentrated worlds*" such (legal) networks [3].

## 4. ACKNOWLEDGMENTS

R. Boulet benefits from a post doctoral grant of the Institut National des Sciences de l'Univers (CNRS, Paris). The legal data are built by the Dept. *Recht, Regierung, Technologien* (Centre Marc Bloch, CNRS, Berlin). Many thanks to E. Catta (*French rapporteur*, *Commission Supérieure de Codification*) for her many useful advices. This study is funded by the RTRA STAE in Toulouse (MAELIA project (http://www.iaai-maelia.eu/ ). The yEd Graph editor is used for producing the figure. Statistical properties of networks have been computed with *R* and the library *igraph* (http://www.r-project.org/).

## 5. REFERENCES

1. Boulet, R. 2008. Comparaison de graphes, applications à l'étude d'un réseau de sociabilité paysan au Moyen Âge. Doctoral Thesis ; Université de Toulouse II, Le Mirail.

2. Boulet, R., Jouve, B., Rossi, F., and Villa, N. 2008. Batch kernel SOM and related Laplacian methods for social network analysis. Neurocomputing, 71, 1257-1273.

3. Boulet, R., Mazzega, P. et Bourcier, D. 2009. Analyse d'un graphe juridique dense – Le monde concentré des codes législatifs. Technique et Science Informatiques, n° spécial Graphes de Terrain, M. Latapy ed., *soumis*.

4. Bourcier, D., and Mazzega, P. 2007a. Toward measure of complexity in legal systems. . ICAIL'07, June 4-8, Stanford, California USA, ACM Press, 211-215.

5. Bourcier, D., and Mazzega, P. 2007b. Codification law article and graphs, *Legal Knowledge and Information Systems, Jurix 2007*, A.R. Lodder and L. Mommers (eds.), IOS Press, 29-38.

6. Chung, F., and Lu, L. 2006. Complex Graphs and Networks. American Mathematical Society.

7. Clauset, A., Newman, M.E.J., and Moore, C. 2004. Finding community structure in very large networks. Physical Review E, 70, Issue 6.

8. LEGIFRANCE (2008) Service Public de la Diffusion du Droit, http://www.legifrance.gouv.fr/ .

9. Newman, M. E. J., and Park, J. 2003. Why social networks are different from other types of networks. Physical Review E, 68, 036122.

10. Ost, F., and van de Kerchove, M. 2002.De la pyramide au réseau ? Pour une théorie dialectique du droit. Publ. Faculté Univ. Saint-Louis, Bruxelles, 596 pp.

11. Watts, D. J., and Strogatz, S. H. 1998. Collective dynamics of *'small world'* networks. Nature, 393, 440-442.

12. Zhang, P., and Koppaka, L. 2007. Semantics-based legal citation network. ICAIL'07, June 4-8, Stanford, California USA, ACM Press, 123-130.